\documentclass{article}

\PassOptionsToPackage{numbers, compress}{natbib}

\usepackage[final]{neurips_fitml_2024}

\usepackage[utf8]{inputenc} 
\usepackage[T1]{fontenc}    
\usepackage{hyperref}       
\usepackage{url}            
\usepackage{booktabs}       
\usepackage{amsfonts}       
\usepackage{nicefrac}       
\usepackage{microtype}      
\usepackage[table]{xcolor}         

\usepackage[utf8]{inputenc} 
\usepackage[T1]{fontenc}    
\usepackage{microtype}      
\usepackage{nicefrac}       
\usepackage{enumitem}

\usepackage{titletoc}
\usepackage[page,header]{appendix}

\usepackage{url}            
\usepackage{hyperref}
\hypersetup{
    colorlinks,
    citecolor=blue,
    filecolor=blue,
    linkcolor=blue,
    urlcolor=blue,
}

\usepackage{graphicx}
\usepackage{chngcntr} 
\begingroup
	\expandafter\ifx\csname pdfsuppresswarningpagegroup\endcsname\relax\else\global\pdfsuppresswarningpagegroup=1\relax\fi
\endgroup

\usepackage{makecell}
\usepackage{tabularx}
\usepackage{wrapfig}
\usepackage{booktabs}
\usepackage{multirow}
\usepackage{colortbl}
\usepackage{tablefootnote}
\usepackage{array}
\newcolumntype{P}[1]{>{\centering\arraybackslash}p{#1}}
\usepackage{tikz}

\usepackage{amsmath,amssymb,amsfonts,amsthm}
\usepackage{scalerel}
\usepackage{bm}

\usepackage{caption}
\usepackage{subcaption}
\usepackage{algorithm,algcompatible}
\usepackage{algpseudocode}

\definecolor{bright-gray}{HTML}{EEEEEE}
\DeclareMathOperator{\Tr}{Tr}

\newcommand{\baseline}{MAP + GS}
\newcommand{\baselineLong}{MAP + grid search (GS)}

\title{Learning the Regularization Strength for Deep Fine-Tuning via a Data-Emphasized\\Variational Objective}

\author{%
  Ethan Harvey\thanks{Both authors made lead author level contributions} \\
  Department of Computer Science\\
  Tufts University\\
  \texttt{ethan.harvey@tufts.edu} \\
  \And
  Mikhail Petrov\footnotemark[1] \\
  Department of Mechanical Engineering\\
  Tufts University\\
  \texttt{mikhail.petrov@tufts.edu} \\
  \And
  Michael C. Hughes \\
  Department of Computer Science\\
  Tufts University\\
  \texttt{michael.hughes@tufts.edu} \\
}

\begin{document}

\maketitle

\begin{abstract}
  A number of popular transfer learning methods rely on grid search to select regularization hyperparameters that control over-fitting. 
  This grid search requirement has several key disadvantages: the search is computationally expensive, requires carving out a validation set that reduces the size of available data for model training, and requires practitioners to specify candidate values.
  In this paper, we propose an alternative to grid search: directly learning regularization hyperparameters on the full training set via model selection techniques based on the \emph{evidence lower bound} (``ELBo'') objective from variational methods.
  For deep neural networks with millions of parameters, we specifically recommend a modified ELBo that upweights the influence of the data likelihood relative to the prior while remaining a valid bound on the evidence for Bayesian model selection.
  Our proposed technique overcomes all three disadvantages of grid search. We demonstrate effectiveness on image classification tasks on several datasets, yielding heldout accuracy comparable to existing approaches with far less compute time.
\end{abstract}

\section{Introduction}

\let\thefootnote\relax\footnotetext{Open-source code: \url{https://github.com/tufts-ml/data-emphasized-ELBo}}

When fine-tuning deep neural networks (DNNs), a significant amount of computational resources are devoted to tuning hyperparameters that control model complexity to manage tradeoffs between under- and over-fitting on the target task of interest. One widespread example would be tuning the value of the scalar multiplier that controls the strength of an additive loss term computed as the sum-of-squares on weight coefficient values, known in various communities as L2 regularization \citep{murphy2022Regularization}, Ridge penalty \citep{hastie2009elements,kobak2020optimal}, or ``weight decay'' \citep{krogh1991simple,goodfellow2016Regularization}. 
A common technique for tuning such hyperparameters is to hold out a dedicated validation set and use grid search to find the hyperparameters that perform best on the validation set \citep{raschka2018model,murphy2022Picking}.

While reasonably effective and in widespread use to manage over-fitting in recent transfer learning \citep{xuhong2018explicit,shwartz2022pre}, using grid search for hyperparameter selection has three key drawbacks. First and perhaps most important, the need to train separate models at each possible value in the grid significantly increases computational runtime and resources.
Second, the need to carve out a validation set to assess performance reduces the amount of available data that can inform model training. This can cause under-fitting, especially when available data has limited size.
Finally, grid search requires a list of candidate values specified in advance, yet ideal values may vary widely depending on the data and specific classification task at hand.

We take another approach to hyperparameter selection, inspired by a pragmatic Bayesian perspective. 
Suppose we model observable dataset $\mathcal{D}$ via a likelihood $p( \mathcal{D} | \theta)$, where $\theta$ is a high-dimensional parameter to be estimated, with prior distribution $p( \theta | \eta)$ controlled by hyperparameter $\eta$ (say just 1-5 dimensions). Instead of point estimating a specific $\theta,\eta$ pair, we can instead estimate a posterior $p(\theta | \mathcal{D}, \eta)$ while simultaneously directly learning $\eta$ to optimize $p( \mathcal{D} | \eta) = \int_{\theta} p( \mathcal{D}, \theta | \eta) d\theta$. This latter objective $p( \mathcal{D} | \eta)$ is known as the \emph{marginal likelihood} or \emph{evidence} \cite{lotfi2022bayesian}. The evidence naturally encodes a notion of Occam’s razor, favoring the hyperparameter setting that leads to the simplest model that fits the data well, while penalizing complex models that over-fit the training data \cite{jeffreys1939theory,mackay1991bayesian,bishop2006Bayesian}.
Learning $\eta$ to maximize evidence (or equivalently, the logarithm of evidence) via gradient descent avoids all three issues with grid search: we need only one run of gradient descent (not separate efforts for each candidate $\eta$ value in a grid), we can use all available labeled data for training without any validation set, and we can explore the full continuous range of possible $\eta$ values rather than a limited discrete set that must be predefined.

While elegant in theory, this vision of selecting hyperparameters via maximizing evidence is difficult in practice for most models of interest due to the intractable high-dimensional integral that defines the evidence. For modern deep image classifiers with millions of parameters, computing the evidence directly seems insurmountable even for a specifc $\eta$, let alone optimizing evidence to select a preferred $\eta$ value. In this work, we use and extend tools from variational Bayesian methods \citep{blei2017variational,jordan1999introduction}, specifically tractable lower bounds on the evidence,  to make hyperparameter selection for fine-tuning deep neural image classifiers possible. 

Ultimately, we contribute methods that should help practitioners perform cost-effective transfer learning on custom datasets. When available data is plentiful, our experiments suggest our approach is competitive in accuracy while reducing total training time from 16 hours for L2-SP \cite{xuhong2018explicit} and 149 hours for PTYL \cite{shwartz2022pre} (using the grid search ranges recommended by the original authors) to under 3 hours.
When available data is limited, e.g., only 5-300 labeled examples per class, our experiments suggest our approach can be particularly effective in improving accuracy and runtime.

\section{Background}

\textbf{Problem setup.}
Consider training a neural network classifier composed of two parts. First, we have a backbone encoder $f$ with weights $w \in \mathbb{R}^D$, which non-linearly maps input vector $x_i$ to a representation vector $z_i \in \mathbb{R}^H$ (which includes an ``always one'' feature to handle the need for a bias/intercept). The second part is a linear-decision-boundary classifier head with weights $V \in \mathbb{R}^{C \times H}$, which leads to probabilistic predictions over $C$ possible classes. We wish to find values of these parameters that produce good classification decisions on a provided \emph{target task} dataset of $N$ pairs $x_i, y_i$ of features $x_i$ and corresponding class labels $y_i \in \{1, 2, \ldots C\}$. For transfer learning, we assume the backbone weights $w$ have been pretrained to high-quality values $\mu$ on a source task.

\textbf{Deep learning view.}
Typical approaches to transfer learning in the deep learning tradition (e.g., baselines in \cite{xuhong2018explicit}) would pursue empirical risk minimization with an L2-penalty on weight magnitudes for regularization, training to minimize the loss function
\begin{align}
    L(w, V) := \frac{1}{N} \left( \sum_{i=1}^N \ell(y_i, V f_w( x_i) ) + \frac{\alpha}{2} || w ||_2 ^2 + \frac{\beta}{2} || \text{vec}(V) ||_2 ^2 \right),
    \label{eq:loss_with_l2_penalty}
\end{align}
where $\ell$ represents a cross-entropy loss indicating agreement with the true label $y_i$ (one of $C$ categories), while the L2-penalty on weights $w,V$ encourages their magnitude to \emph{decay} toward zero, and this regularization is thus often referred to as ``weight decay''.
The key hyperparameters $\alpha \geq 0$, $\beta \geq 0$ encode the strength of the L2 penalty, with higher values yielding simpler representations and simpler decision boundaries.
Model training would thus consist of solving $w^*, V^* \gets \arg\!\min_{w,V} L(w,V)$ via stochastic gradient descent, given fixed hyperparameters $\alpha, \beta$. 
In turn, the values of $\alpha,\beta$ would be selected via grid search seeking to optimize $\ell$ or error on a validation set. 

\textbf{Bayesian view.} Bayesian interpretation of this neural classification problem would define a joint probabilistic model $p(w, V, y_{1:N})$ decomposed into factors $p(w)p(V) \prod_{i} p(y_i | w, V)$ defined as:
\begin{align}
\label{eq:joint_pdf_model}
    p(w) &{=} \mathcal{N}( w | \mu_p, \lambda \Sigma_p),
    ~~p(V) {=} \mathcal{N}( \text{vec}(V) | 0, \tau I),
    ~~p(y_i | w, V) {=} \text{CatPMF}( y_i | \textsc{sm}( V f_w(x_i) ) ).
\end{align}
Here, $\lambda \geq 0, \tau \geq 0$ are  hyperparameters controlling over/under-fitting, $\mu_p, \Sigma_p$ represent \emph{a priori} knowledge of the mean and covariance of backbone weights $w$ (see paragraph below), and $\textsc{sm}$ is the softmax function. Note $x_i$ is a known fixed value, not a random variable in the model; we leave such fixed quantities out of probabilistic conditioning notation for simplicity.
To fit this model, pursing maximum a-posteriori (MAP) estimation of both $w$ and $V$ recovers the objective in Eq.~\eqref{eq:loss_with_l2_penalty} when we set
$\alpha = \frac{1}{\lambda}$, $\beta = \frac{1}{\tau}$, $\mu_p=0, \Sigma_p=I$ and have $\ell$ set to $- \log p(y_i | w, V)$.

\textbf{Need for validation set and grid search.}
Selecting $\alpha,\beta$ (or equivalently $\lambda,\tau$) to directly minimize Eq.~\eqref{eq:loss_with_l2_penalty} on the training set alone is not a coherent way to guard against over-fitting. Regardless of data content or weight parameter values, we would select $\alpha^* = 0,\beta^* = 0$ to minimize $L$ as a function of $\alpha,\beta$ and thus enforce no penalty on weight magnitudes at all. Carving out a validation set for selecting these hyperparameters is thus critical to avoid over-fitting when point estimating $w,V$.

\setlength{\tabcolsep}{4pt}
\begin{wraptable}{r}{0.35\textwidth}
  \caption{Mean and covariance of backbone weights $w$ for several transfer learning approaches.}
  \label{tab:transfer_learning_methods}
  \centering
  \footnotesize
  \begin{tabular}{lcc}
    \toprule \bfseries Method & \bfseries $p(w)$ & Init. \\ \midrule
    L2-zero & $\mathcal{N}(0, \lambda I)$ & $\mu$ \\
    L2-SP & $\mathcal{N}(\mu, \lambda I)$ & $\mu$ \\
    PTYL & $\mathcal{N}(\mu, \lambda \Sigma)$ & $\mu$ \\
    \bottomrule
  \end{tabular}
\end{wraptable}
\setlength{\tabcolsep}{6pt}

\paragraph{Backbone prior mean/covariance.}
Several recent transfer learning approaches correspond to specific settings of the backbone mean and covariance $\mu_p, \Sigma_p$. Let vector $\mu$ represent a specific setting of pretrained backbone weights $w$ that performs well on a source task.
Setting $\mu_p{=}0, \Sigma_p{=}I$ recovers the conventional approach to transfer learning, which we call L2-zero, where regularization pushes backbone weights to zero and the pretrained value $\mu$ only informs the initial value of backbone weights $w$ before any SGD~\citep{xuhong2018explicit,harvey2024transfer}.  Instead, setting the prior mean as $\mu_p = \mu$ along with $\Sigma_p = I$ recovers the \emph{L2 starting point} (L2-SP) regularization method of \citep{xuhong2018explicit}, also covered in \citep{chelba2006adaptation}. Further setting $\Sigma_p$ to the estimated covariance matrix $\Sigma$ of a Gaussian approximation of the posterior over backbone weights for the source task recovers ``Pre-Train Your Loss'' (PTYL)~\citep{shwartz2022pre}.

\textbf{Need to specify a search space.}
Selecting $\alpha,\beta$ (or equivalently $\lambda,\tau$) via grid search requires practitioners to specify a grid of candidate values spanning a finite range.
For the PTYL method, the optimal search space for these key hyperparameters is still unclear.
For the same prior and the same datasets, the search space has varied between works: originally, the method creators recommended large values from 1e0 to 1e10 \cite{shwartz2022pre, harvey2024transfer}. Later works search far smaller values (1e-5 to 1e-3) \cite{rudner2024finetuning}.

\section{Methods}
\label{sec:methods}

\subsection{Variational methods for posterior estimation}
\newcommand{\JELBO}{J_{\text{ELBo}}}

For most complex probabilistic models, such as $p(y_{1:N}, w, V)$ defined in Eq.~\eqref{eq:joint_pdf_model}, evaluating the posterior distribution probability density function (PDF) $p(w, V | y_{1:N})$ is challenging even for specific $w, V$ value, because $p(w, V | y_{1:N}) \propto \frac{1}{p(y_{1:N})} p( y_{1:N}, w, V)$, and the normalization term is exactly the intractable evidence $p(y_{1:N})$, defined as a challenging high-dimensional integral $p(y_{1:N}) = \int p( y_{1:N}, w, V) dw dV$. This evaluation roadblock also makes sampling from the posterior difficult. As a remedy, variational methods \cite{blei2017variational} provide a framework to estimate an approximate posterior $q(w, V) \approx p( w, V | y_{1:N})$ that belongs to a simpler parametric family of distributions. For example, we may define $q$ as
\begin{align}
    q(w, V) = q(w) q(V), 
    q(w) {=} \mathcal{N}( w | \bar{w}, \bar{\sigma}^2 I),
    q(V) {=} \mathcal{N}( \text{vec}(V) | \text{vec}(\bar{V}), \bar{\sigma}^2 I).
\end{align}
We denote the free parameters of $q$ with bars to distinguish them from model parameters. Weights $\bar{w}, \bar{V}$ represent means of the backbone and classifier head. Scalar $\bar{\sigma}^2 > 0$ controls variance around these means. We use one variance parameter for simplicity; future work could allow separate parameters for different classifier parts. 
Well-known results in variational inference then suggest fitting the parameters of $q$ by maximizing an objective function known as the \emph{evidence lower bound} (``ELBo'', \citep{blei2017variational}), denoted as $J_{\text{ELBo}}$ and defined for our model $p$ and approximate posterior $q$ as:
\begin{align}
    \JELBO := \mathbb{E}_{q(w,V)} \left[ \sum_{i=1}^{N} \log p(y_i | w, V) \right] - \mathbb{KL}(q(w) \| p(w|\lambda))  - \mathbb{KL}(q(V) \| p(V| \tau)).
\label{eq:elbo_objective}
\end{align}
This objective is a function of data $y_{1:N}, x_{1:N}$, variational posterior parameters $\bar{w}, \bar{V}, \bar{\sigma}$, and prior hyperparameters $ \lambda, \tau$. Maximizing $\JELBO$ can be shown equivalent to finding the specific $q$ that is ``closest'' to the true posterior (in the sense of KL divergence). Further, as the name of the objective suggests, we can show mathematically that $\JELBO( y_{1:N}, \ldots, \lambda, \tau ) \leq \log p( y_{1:N} | \lambda, \tau)$. That is, the ELBo is a lower bound on the log evidence (which itself is a function of the chosen $\lambda,\tau$).
This relation suggests a potential utility for data-driven selection of these hyperparameters. Prior work~\citep{cherief2019consistency} argues that using ELBo for model selection has strong theoretical guarantees, even under misspecification.

\paragraph{Optimizing the ELBo for neural net classifiers.}
Of the 3 additive terms defining the ELBo objective, both KL terms are the most tractable to evaluate and compute gradients of, as they involve KL divergences of multivariate Gaussians amenable to closed-form expressions. The first term, the expected log likelihood, requires slightly more care.
To handle this non-conjugate expectation for neural network classifiers a key insight from \citep{blundell2015weight} suggests applying the ``reparameterization trick'', a general way to estimate gradients in Monte Carlo fashion~\citep{xu2019variance,mohamed2020monte}.
To implement this estimator for this likelihood term, we first draw vectors the same size as $w$ and $\text{vec}(V)$, denoted $\epsilon^w$ and $\epsilon^V$, from a standard normal. Then, we transform to samples of $w,V$ from $q$ via well-known transformations (e.g., for the backbone: $w \gets \bar{w} + \bar{\sigma} \epsilon^w$). Averaging over such independent and identically distributed (IID) samples $w,V$ from $q$ allows a Monte Carlo approximation of the expected log likelihood. Gradients follow via automatic differentiation of that Monte Carlo estimator's computational graph. We find using \emph{just one sample} per training step is sufficient and most efficient.

\textbf{Related work on Bayesian neural nets.}
Variational inference for neural networks has a long history, dating back to work by \citet{hinton1993keeping} and \citet{graves2011practical}. Modern efforts include MOdel Priors Extracted from Deterministic DNN (MOPED) \cite{krishnan2019efficient}, which focuses on using informative priors to enable scalable variational inference, not to select regularization hyperparameters.

\subsection{Learning key hyperparameters $\lambda, \tau$ with the ELBo}

While most works on variational inference concentrate on learning just the posterior over model parameters, it is staightforward to optimize the ELBo objective for prior hyperparameters $\lambda, \tau$ as well. Recall that the term $-\mathbb{KL}(q(w) \| p(w))$ is tractable since both the approximate posterior and prior are multivariate Gaussian.
For instance, in our particular model in Eq.~\eqref{eq:joint_pdf_model}, the KL divergence between two Gaussians \citep{murphy2022Example} simplifies for the backbone KL term as:
\begin{align}
    -\mathbb{KL}(q(w) \| p(w)) = -\frac{1}{2} \left[ \frac{\bar{\sigma}^2}{\lambda} \Tr (\Sigma_p^{-1}) + \frac{1}{\lambda} (\mu_p-\bar{w})^T\Sigma_p^{-1}(\mu_p-\bar{w}) - D + \log \left( \frac{\lambda^{D}\det(\Sigma_p)}{\bar{\sigma}^{2D}} \right)\right].
\end{align}
A simpler expression is possible for the KL over the classifier head weights $V$ (see App.~\ref{sec:learning_key_hyperparameters}).

\textbf{Closed-form updates.} To find an optimal $\lambda$ value with respect to the $\JELBO$, notice that of the 3 additive terms in Eq.~\eqref{eq:elbo_objective}, only the KL term between $q(w)$ and $p(w)$ involves $\lambda$. We solve for $\lambda$ by taking the gradient of the KL term with respect to $\lambda$, setting to zero, and solving, with assurances of a local maximum of $\JELBO$ via a second derivative test (see App.~\ref{sec:second_derivative_test}). The gradient is
\begin{align}
    \nabla_\lambda -\mathbb{KL}(q(w) \| p(w)) = -\frac{1}{2} \left[ - \frac{\bar{\sigma}^2}{\lambda^2} \Tr(\Sigma_p^{-1}) - \frac{1}{\lambda^2} (\mu_p-\bar{w})^T \Sigma_p^{-1} (\mu_p-\bar{w}) + \frac{D}{\lambda} \right].
\end{align}
Setting $\nabla_\lambda -\mathbb{KL}(q(w) \| p(w)) = 0$ and solving for $\lambda$, we get 
\begin{align}
    \lambda^* = \frac{1}{D} \Big[ \bar{\sigma}^2 \Tr(\Sigma_p^{-1}) + (\mu_p-\bar{w})^T \Sigma_p^{-1} (\mu_p-\bar{w}) \Big].    
    \label{eq:lambda_update}
\end{align}
Similar updates can be derived for $\tau$ (see App.~\ref{sec:learning_key_hyperparameters}). 

\subsection{Inflating dataset size to select complex models with better generalization}

\begin{figure}[!t]
  \centering
  \includegraphics[width=0.6666\linewidth]{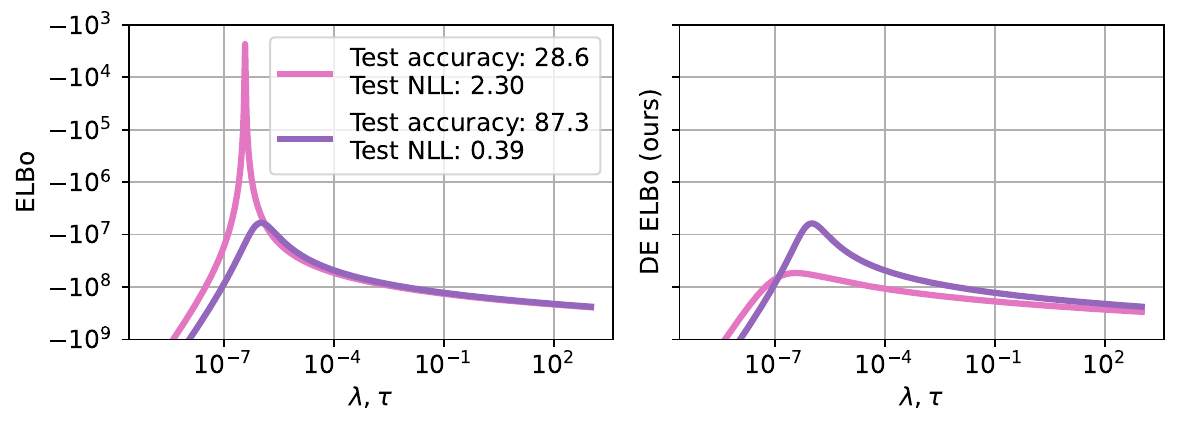}
  \caption{Model selection comparison between the ELBo (left) and our \emph{data-emphasized ELBo} (DE ELBo) (right) for two ResNet-50s trained on CIFAR-10 $N=1000$.
  For both models, we fix the estimated posterior $q$ and vary $\lambda, \tau$.
  \textbf{Takeaway: Without enough training data or with too many model parameters, the ELBo has a preference for simpler models.}}
  \label{fig:elbo_comparison}
\end{figure}

In our intended transfer learning settings with only a few 10s or 100s of labeled examples in available training data, classical statistical learning would hardly recommend deep neural network backbones with millions of parameters (recall $w \in \mathbb{R}^D$, with $D$ very large). 
However, the trend in deep learning has suggested that careful fine-tuning of very large models can deliver strong performance that generalizes well \citep{sharif2014cnn,xuhong2018explicit}.
We find that straightforward use of the ELBo in this regime (when $D \gg N$) is overly conservative in hyperparameter selection, preferring simpler models with much-worse performance at the target classification task (see demo in Fig.~\ref{fig:elbo_comparison}, described further below). When $N$ is much smaller than $D$, the KL terms dominate Eq.~\eqref{eq:elbo_objective}, overpowering the likelihood signal.

In order to make an objective that emphasizes the need for strong data fit in this overparameterized regime, we modify the usual ELBo by introducing a scaling factor $\kappa \geq 1$ on the likelihood term, 
\begin{align}
J_{\text{DE ELBo}} := \kappa \mathbb{E}_{q(w,V)} \left[ \sum_{i=1}^{N} \log p(y_i | w, V) \right] - \mathbb{KL}(q(w, V) \| p(w, V) )
\label{eq:rescaled_elbo_objective}
\end{align}
When $D \gg N$, we recommend setting $\kappa = D/N$ to achieve an improved balance between likelihood and KL terms. In our applications, with $D$ in millions and $N \geq 100$, this yields $\kappa \gg 1$. Of course, $\kappa=1$ recovers the standard ELBo.
We call this objective the \emph{data-emphasized ELBo} (DE ELBo). We use it with $\kappa = D/N$ for all training steps, as well as ultimate hyperparameter/model selection.

\textbf{Justification.} Beyond later empirical success, we offer two arguments suggesting this revised objective as suitable for hyperparameter selection.
One justification is that the function in Eq.~\eqref{eq:rescaled_elbo_objective} maintains a valid lower bound on the evidence $\log p( y_{1:N} )$ of our observed data of size $N$, because as long as each $y_i$ is a discrete random variable (in our case, a 1-of-C class label), the term we are inflating is a log PMF and thus is always negative ($\mathbb{E}_{q}[\log p( y_i | w, V)] \leq 0$). While the bound may be ``loose'' in an absolute sense, what matters more is which $\lambda,\tau$ values are favored.
Another justification views this approach as acting as if we are modeling $\kappa N$ IID data instances, and we just happen to observe $\kappa$ copies of the size-$N$ dataset $y_{1:N}$ with known features $x_{1:N}$.

\textbf{Demonstrating the value of $\kappa=D/N$ for improved selection.}
In Fig.~\ref{fig:elbo_comparison}, we compare the cases of $\kappa=1$ (left panel) and $\kappa=D/N$ (right) on CIFAR-10 for L2-SP. In each plot, we compare two possible $q$, ``A'' (in pink) and ``B'' (in purple), across a range of $\lambda,\tau$ values. Version $A$ sets $\bar{w}_A,\bar{V}_A, \bar{\sigma}_A$ to a solution favored by ELBo, with known test-set accuracy 28.6\%. Version $B$ sets $\bar{w}_B,\bar{V}_B, \bar{\sigma}_B$ to values favored by our \emph{data-emphasized ELBo}, with known test-set accuracy of 87.3\%. We see the poor-accuracy version A model is strongly favored by the conventional ELBo, while the more accurate version B model is favored by our recommended DE ELBo with $\kappa=D/N$. 

\textbf{Related work adjusting Bayesian objectives.}
Other approaches have recognized value in raising the likelihood to a power in the context of general Bayesian modeling, under the vocabulary terms of a tempered or \emph{power likelihood} \citep{antoniano2013bayesian}, \emph{power posterior} \citep{friel2008marginal,miller2019robust}, or ``Safe'' Bayesian learning~\citep{grunwald2012safe,grunwald2017inconsistency}. 
However, throughout \emph{all} these previously cited cases, the desired power is meant to be \emph{smaller than one}, with the stated purpose of counter-acting misspecification. Values of $\kappa$ larger than one (amplifying influence of data) are not even considered in these previous works, and their applications are far from our focus on transfer learning for deep neural networks. Others have recommended upweighting the KL term (not the likelihood), as in $\beta$-variational autoencoders~\citep{higginsBetaVAE2017}, again using numerical values that diminish rather than emphasize the likelihood.

Bayesian Data Reweighting~\citep{wang2017robust} learns instance-specific likelihood weights, some of which can be larger than one. However, its primary motivation is robustness and the ability to turn down the influence of observations that do not match assumptions. The authors discourage letting observations be ``arbitrarily up- or down-weighted''. In work with similar spirit to ours, Power sLDA artificially inflates the likelihood of class labels relative to words in supervised topic modeling applications~\citep{zhang2015supervise}. However, they did not pursue hyperparameter selection or DNNs, as we do here, instead they focused on different models with only a few thousand parameters. Work in Bayesian deep learning has used multipliers to adjust the ``temperature'' of the entire log posterior (not just the likelihood), in a line of work known as \emph{cold posteriors} \cite{wenzel2020good, kapoor2022uncertainty}.

\subsection{Overall algorithm and implementation}
Ultimately, our fitting algorithm proceeds by doing stochastic gradient descent (SGD) for $\bar{w}, \bar{V}$, and $\bar{\sigma}$.
We parameterize the standard deviation $\bar{\sigma} = \log(1+\exp(\rho))$ with free parameter $\rho \in \mathbb{R}$, which ensures $\bar{\sigma}(\rho)$ is always positive during gradient descent.
Before each gradient-based update step to these parameters of $q$, we perform closed-form updates of $\lambda,\tau$ as in Eq.~\eqref{eq:lambda_update}; this update is the same regardless of $\kappa$. We run until a specified maximum number of iterations is reached.

Ultimately, our proposed ELBo-based method can deliver an estimated posterior $q$ and learned hyperparameters $\lambda,\tau$ from one run of gradient descent. 
We compare to a baseline that simply performs MAP point-estimation of $w,V$, with a separate SGD run at each candidate $\lambda, \tau$ configuration in a fixed grid (see App.~\ref{sec:classifier_details}). This ``grid search'' baseline is representative of cutting-edge work in transfer learning~\citep{shwartz2022pre,harvey2024transfer}.
In all experiments, we select ResNet-50 \cite{he2016deep} as the backbone $f_w$. 

Each run of our method and the baseline depends on the adequate selection of learning rate. All runs search over 4 candidate values and select the best according to either the DE ELBo (ours) or validation-set likelihood (baseline).

Both our method and the baseline can be implemented with any of the 3 settings of the backbone prior in Tab.~\ref{tab:transfer_learning_methods}. For the PTYL method \citep{shwartz2022pre}, we use released code from \citet{harvey2024transfer}, which fixes a key issue in the original implementation so that the learned low-rank covariance $\Sigma_p$ is properly scaled. We use the Woodbury matrix identity \citep{woodbury1950inverting}, trace properties, and the matrix determinant lemma to compute the trace of the inverse, squared Mahalanobis distance, and log determinant of low-rank covariance matrix $\Sigma_p$ for the KL term.
See App.~\ref{sec:low-rank} for details.

\section{Results}

Across several probabilistic modeling priors for transfer learning and 2 datasets, our findings are:

\paragraph{Our \emph{data-emphasized ELBo} makes learning the regularization strength possible without compromising task accuracy.}
As demonstrated in Fig.~\ref{fig:elbo_comparison}, in regimes where the parameter dimension $D$ is much larger than $N$, the ELBo has a preference for simpler models favored by the prior, potentially at the expense of downstream task accuracy. By setting $\kappa=D/N$, we upweight the likelihood terms to have reasonable importance relative to the KL terms on the ELBo. This results in an objective that can simultaneously learn backbone and classifier head weights $\bar{w},\bar{V}$ and key hyperparameters $\lambda, \tau$ that are effective at downstream tasks.

\begin{figure}[t!]
  \centering
  \includegraphics[width=\linewidth]{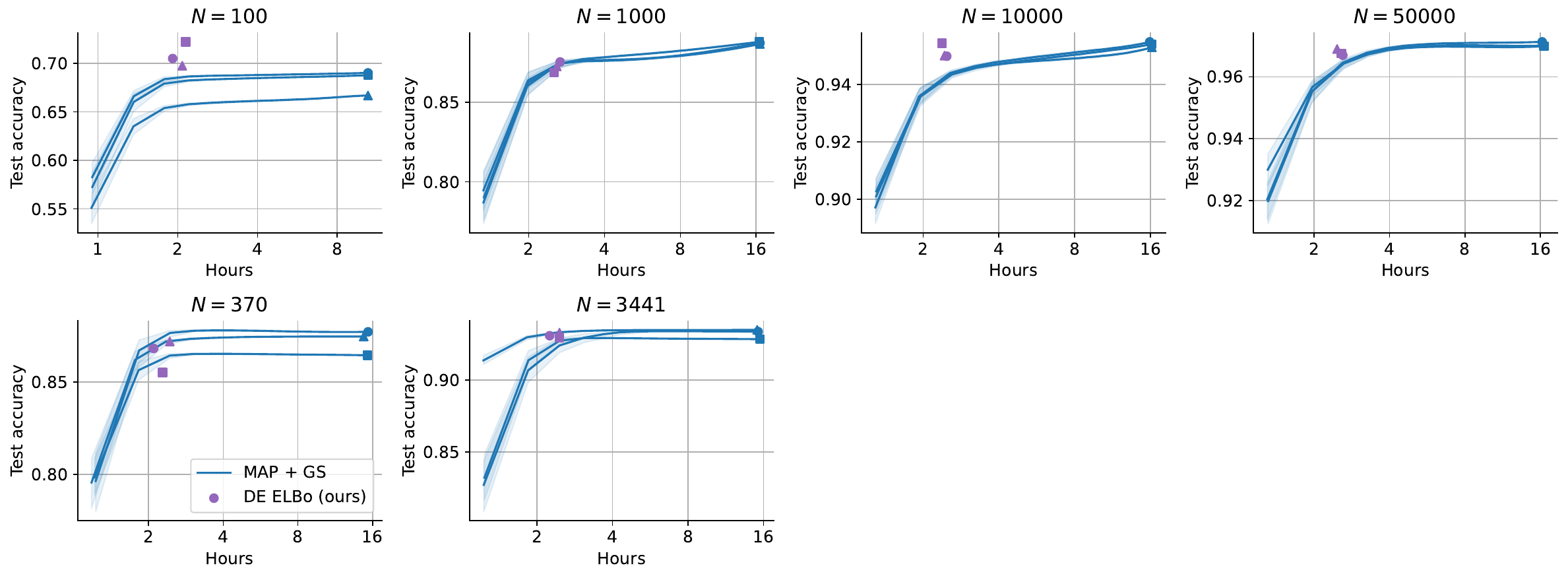}
  \caption{Test-set accuracy on CIFAR-10 (top row) and Oxford-IIIT Pet (bottom row) over training time for L2-SP with \emph{\baselineLong} and our \emph{data-emphasized ELBo} (DE ELBo). We run each method on 3 separate training sets of size $N$ (3 different marker styles).
  \textbf{Takeaway: Our DE ELBo achieves as good or better performance at small dataset sizes and similar performance at large dataset sizes with far less compute time.} To make the blue curves, we did the full grid search once (markers). Then, at each given shorter compute time, we subsampled a fraction of all hyperparameter configurations with that runtime and chose the best via validation NLL. Averaging this over 500 subsamples at each runtime created each blue line.
  }
  \label{fig:computational_time_comparison}
\end{figure}

\textbf{The runtime of our DE ELBO is affordable, and avoids the extreme runtime costs of grid search.}
In Tab.~\ref{tab:computational_time_comparison}, we show that an individual SGD run of our DE ELBO has comparable cost to one SGD run of standard MAP estimation. However, the cumulative cost of grid search needed to select $\lambda,\tau$ for the MAP baseline is far higher than our approach: for PTYL the recommended grid search costs over 149 hours; our approach delivers in under 3 hours.

\paragraph{Our \emph{data-emphasized ELBo} achieves heldout accuracy comparable to existing approaches with far less compute time.}
In Fig.~\ref{fig:computational_time_comparison}, we compare accuracy on CIFAR-10 \citep{krizhevsky2010cifar} and Oxford-IIIT Pet \citep{parkhi2012cats} test sets over training time for L2-SP with \baseline\ and our DE ELBo.
Our DE ELBo achieves comparable heldout accuracy with far less compute time.
When training data is limited in size (CIFAR-10 at $N=100$ in Fig.~\ref{fig:computational_time_comparison}), our approach can perform even better than grid search.

In Tab.~\ref{tab:CIFAR-10_acc} and \ref{tab:Oxford-IIIT_Pet_acc}, we report the final accuracy on CIFAR-10 and Oxford-IIIT Pet test sets using the entire grid search compared to our DE ELBo. See App.~\ref{sec:nll_results} for NLL results.

\begin{table}[htbp!]
  \caption{Computational time comparison between methods for transfer learning with informative priors using grid search to find the hyperparameters that perform best on the validation set then retraining with the selected hyperparameters on the combined set of all $N$ images (merged train and validation) and using our \emph{data-emphasized ELBo} (DE ELBo) to learn $\lambda, \tau$ and select the initial learning rate (lr) on the training set for CIFAR-10 at $N=50000$. See App.~\ref{sec:classifier_details} for search space details. Runtime measured on one NVIDIA A100 40GB PCIe GPU.}
  \label{tab:computational_time_comparison}
  \centering
  \footnotesize
  \begin{tabular}{llcccc}
    \toprule
    \bfseries Model & \bfseries Method & \bfseries Avg. SGD runtime & \bfseries lr search space & \bfseries $\lambda,\tau$ search space & \bfseries Total GS time \\
    \midrule    
    \rowcolor{bright-gray} L2-SP & \baseline & 39 mins. 11 secs. & 4 & 6 & \phantom{0}16 hrs. 15 mins. \\ 
    & DE ELBo & 39 mins. \phantom{0}0 secs. & 4 & n/a & \phantom{00}2 hrs. 36 mins. \\ 
    \rowcolor{bright-gray} PTYL & \baseline & 37 mins. 15 secs. & 4 & 60 & 149 hrs. 36 mins. \\ 
    & DE ELBo & 40 mins. \phantom{0}0 secs. & 4 &  n/a &\phantom{00}2 hrs. 39 mins. \\ 
    \bottomrule
  \end{tabular}
\end{table}
\setlength{\tabcolsep}{6pt}

\setlength{\tabcolsep}{4pt}
\begin{table}[htbp!]
  \caption{Accuracy on CIFAR-10 test set for different probabilistic models and methods. We report mean (min-max) over 3 separately-sampled training sets. For each separately-sampled training set and training set size, the \emph{\baselineLong} baseline requires 24 different SGD runs for L2-zero and L2-SP, and 240 for PTYL. Our \emph{data-emphasized ELBo} (DE ELBo) requires 4 different SGD runs (one for each initial learning rate) and learns optimal $\lambda,\tau$ values. See App.~\ref{sec:classifier_details} for hyperparameter search space details.
  }
  \label{tab:CIFAR-10_acc}
  \centering
  \footnotesize
  \begin{tabular}{llcccc}
    \toprule
    \bfseries Model & \bfseries Method & $N =$ {\bfseries 100 (10/cl.)} & \bfseries 1000 (100/cl.) & \bfseries 10000 (1k/cl.) & \bfseries 50000 (5k/cl.) \\
    \midrule
    \rowcolor{bright-gray} L2-zero & \baseline & 67.7 (66.0-68.6) & 87.8 (87.5-88.4) & 95.0 (94.4-95.5) & 97.2 (97.1-97.2) \\
    & DE ELBo & 60.9 (58.9-63.1) & 87.2 (87.0-87.4) & 91.2 (90.7-92.0) & 93.2 (93.0-93.3) \\
    \rowcolor{bright-gray} L2-SP & \baseline & 68.1 (66.7-68.9) & 87.3 (87.2-87.3) & 95.3 (95.1-95.7) & 97.1 (97.0-97.1) \\
     & DE ELBo & 70.6 (68.7-72.7) & 87.2 (86.8-87.4) & 95.0 (94.8-95.2) & 96.8 (96.7-96.9) \\
    \rowcolor{bright-gray} PTYL & \baseline & 67.5 (65.7-68.4) & 87.9 (86.9-89.2) & 95.2 (95.0-95.4) & 97.3 (97.3-97.3) \\
    & DE ELBo & 70.6 (68.7-72.6) & 87.2 (86.9-87.6) & 95.1 (94.9-95.4) & 96.9 (96.8-96.9) \\
    \bottomrule
  \end{tabular}
\end{table}
\setlength{\tabcolsep}{6pt}

\setlength{\tabcolsep}{4pt}
\begin{table}[htbp!]
  \caption{Accuracy on Oxford-IIIT Pet test set for different probabilistic models and methods. We report mean (min-max) over 3 separately-sampled training sets. For each separately-sampled training set and training set size, the \emph{\baselineLong} baseline requires 24 different SGD runs for L2-zero and L2-SP, and 240 for PTYL. Our \emph{data-emphasized ELBo} (DE ELBo) requires 4 different SGD runs (one for each initial learning rate) and learns optimal $\lambda,\tau$ values. See App.~\ref{sec:classifier_details} for hyperparameter search space details.}
  \label{tab:Oxford-IIIT_Pet_acc}
  \centering
  \footnotesize
  \begin{tabular}{llccc}
    \toprule
    \bfseries Model & \bfseries Method & $N~{=}$ {\bfseries 370 (10/cl.)} & \bfseries 3441(93/cl.) \\
    \midrule
    \rowcolor{bright-gray} L2-zero & \baseline & 87.2 (86.6-87.5) & 93.2 (93.0-93.4) \\
    & DE ELBo & 81.8 (80.3-83.6) & 92.6 (92.1-92.4) \\
    \rowcolor{bright-gray} L2-SP & \baseline & 87.2 (86.6-87.6) & 93.2 (93.0-93.4) \\
    & DE ELBo & 86.6 (85.7-87.3) & 93.2 (93.0-93.3) \\
    \rowcolor{bright-gray} PTYL & \baseline & 87.1 (86.6-87.5) & 92.2 (90.7-93.2) \\ 
    & DE ELBo & 86.6 (85.7-87.3) & 93.2 (93.0-93.3) \\
    \bottomrule
  \end{tabular}
\end{table}
\setlength{\tabcolsep}{6pt}

\section{Discussion and Conclusion}

We proposed an alternative to grid search: directly learning regularization hyperparameters on the full training set via model selection techniques based on the ELBo.
We showed that a modified ELBo that upweights the influence of the data likelihood relative to the prior improves model selection with limited training data or an overparameterized model.
We included results on CIFAR-10 and Oxford-IIIT Pet at several data sizes that showed our DE ELBo achieves heldout accuracy comparable to existing approaches with far less compute time.

\paragraph{Learning the regularization strength lets practitioners focus on other aspects that could improve accuracy more.}
For Oxford-IIIT Pet at $N=370$, we found that L2-zero with an initialization pre-trained with supervised learning on ImageNet resulted in a gain of 32.4 percentage points over its self-supervised counterpart (see App.~\ref{sec:supervised_or_self-supervised_priors}).
Our \emph{data-emphasized ELBo} reduces compute by learning the regularization strength which enables practitioners to focus on finding pre-trained weights that generalize better and using data augmentation strategies to improve performance.

\paragraph{What informative prior should I use for PTYL?}
When using PTYL, we found that $\mu, \Sigma$ values obtained from supervised pre-training (minimizing cross-entropy on the source task) lead to better transfer learning (better target task accuracy) than using SimCLR \cite{chen2020simple} self-supervised learning on the source task.
Consistent with \citet{spendl2023easy}'s findings on the 102 Category Flower dataset \citep{nilsback2008automated} without hyperparameter tuning, we find that \citet{shwartz2022pre}'s supervised prior performs better than their self-supervised prior on CIFAR-10 and Oxford-IIIT Pet (see App.~\ref{sec:supervised_or_self-supervised_priors}).

\paragraph{Limitations.}
We acknowledge our proposed approach still requires a (much smaller scale) grid search to select a learning rate.
Additionally, our experiments were limited to two datasets and one backbone architecture, and only look at transfer learning rather than other possible classifier tasks. We hope that more comprehensive experiments in future work can further validate our approach.

\paragraph{Outlook.}
Our proposed approach saves practitioners time by learning an optimal regularization strength without need for expensive grid search. We hope our data-emphasized ELBo for efficient hyperparameter tuning may eventually prove useful across a wide array of classifier tasks beyond transfer learning, such as semi-supervised learning, few-shot learning, continual learning, and beyond.

\subsection*{Acknowledgments}
Authors EH and MCH gratefully acknowledge support in part from the Alzheimer’s Drug Discovery Foundation and the National Institutes of Health (grant \# 1R01NS134859-01). MCH is also supported in part by the U.S. National Science Foundation (NSF) via grant IIS \# 2338962. We are thankful for computing infrastructure support provided by Research Technology Services at Tufts University, with hardware funded in part by NSF award OAC CC* \# 2018149.

\bibliographystyle{plainnat}
\bibliography{neurips_2024}

\begin{thebibliography}{46}
\providecommand{\natexlab}[1]{#1}
\providecommand{\url}[1]{\texttt{#1}}
\expandafter\ifx\csname urlstyle\endcsname\relax
  \providecommand{\doi}[1]{doi: #1}\else
  \providecommand{\doi}{doi: \begingroup \urlstyle{rm}\Url}\fi

\bibitem[Antoniano-Villalobos and Walker(2013)]{antoniano2013bayesian}
Isadora Antoniano-Villalobos and Stephen~G. Walker.
\newblock {Bayesian Nonparametric Inference for the Power Likelihood}.
\newblock \emph{Journal of Computational and Graphical Statistics}, 22\penalty0 (4):\penalty0 801--813, 2013.
\newblock URL \url{http://www.tandfonline.com/doi/abs/10.1080/10618600.2012.728511}.

\bibitem[Bishop(2006)]{bishop2006Bayesian}
Christopher~M. Bishop.
\newblock \emph{Pattern Recognition and Machine Learning}, chapter 3.4 Bayesian Model Comparison.
\newblock Springer, 2006.

\bibitem[Blei et~al.(2017)Blei, Kucukelbir, and McAuliffe]{blei2017variational}
David~M. Blei, Alp Kucukelbir, and Jon~D. McAuliffe.
\newblock {Variational Inference: A Review for Statisticians}.
\newblock \emph{Journal of the American Statistical Association}, 112\penalty0 (518):\penalty0 859--877, 2017.

\bibitem[Blundell et~al.(2015)Blundell, Cornebise, Kavukcuoglu, and Wierstra]{blundell2015weight}
Charles Blundell, Julien Cornebise, Koray Kavukcuoglu, and Daan Wierstra.
\newblock {Weight Uncertainty in Neural Networks}.
\newblock In \emph{International Conference on Machine Learning (ICML)}, 2015.

\bibitem[Chelba and Acero(2006)]{chelba2006adaptation}
Ciprian Chelba and Alex Acero.
\newblock {Adaptation of Maximum Entropy Capitalizer: Little Data Can Help a Lot}.
\newblock \emph{Computer Speech \& Language}, 20\penalty0 (4):\penalty0 382--399, 2006.

\bibitem[Chen et~al.(2020)Chen, Kornblith, Norouzi, and Hinton]{chen2020simple}
Ting Chen, Simon Kornblith, Mohammad Norouzi, and Geoffrey Hinton.
\newblock {A Simple Framework for Contrastive Learning of Visual Representations}.
\newblock In \emph{International Conference on Machine Learning (ICML)}, 2020.

\bibitem[Cherief-Abdellatif(2019)]{cherief2019consistency}
Badr-Eddine Cherief-Abdellatif.
\newblock {Consistency of ELBO maximization for model selection}.
\newblock In \emph{Proceedings of The 1st Symposium on Advances in Approximate Bayesian Inference}, 2019.
\newblock URL \url{https://proceedings.mlr.press/v96/cherief-abdellatif19a.html}.

\bibitem[Friel and Pettitt(2008)]{friel2008marginal}
Nial Friel and Anthony~N. Pettitt.
\newblock {Marginal Likelihood Estimation via Power Posteriors}.
\newblock \emph{Journal of the Royal Statistical Society Series B: Statistical Methodology}, 70\penalty0 (3):\penalty0 589--607, 2008.
\newblock URL \url{https://academic.oup.com/jrsssb/article/70/3/589/7109555}.

\bibitem[Goodfellow et~al.(2016)Goodfellow, Bengio, and Courville]{goodfellow2016Regularization}
Ian Goodfellow, Yoshua Bengio, and Aaron Courville.
\newblock \emph{Deep Learning}, chapter 7 Regularization for Deep Learning.
\newblock MIT Press, 2016.
\newblock URL \url{https://www.deeplearningbook.org/}.

\bibitem[Graves(2011)]{graves2011practical}
Alex Graves.
\newblock Practical variational inference for neural networks.
\newblock In \emph{Advances in Neural Information Processing Systems (NeurIPS)}, 2011.

\bibitem[Gr{\"u}nwald(2012)]{grunwald2012safe}
Peter Gr{\"u}nwald.
\newblock {The Safe Bayesian: learning the learning rate via the mixability gap}.
\newblock In \emph{International Conference on Algorithmic Learning Theory}, pages 169--183. Springer, 2012.

\bibitem[Gr{\"u}nwald and van Ommen(2017)]{grunwald2017inconsistency}
Peter Gr{\"u}nwald and Thijs van Ommen.
\newblock {Inconsistency of Bayesian Inference for Misspecified Linear Models, and a Proposal for Repairing It}.
\newblock \emph{Bayesian Analysis}, 12\penalty0 (4):\penalty0 1069--1103, 2017.
\newblock URL \url{https://projecteuclid.org/journals/bayesian-analysis/volume-12/issue-4/Inconsistency-of-Bayesian-Inference-for-Misspecified-Linear-Models-and-a/10.1214/17-BA1085.full}.

\bibitem[Harvey et~al.(2024)Harvey, Petrov, and Hughes]{harvey2024transfer}
Ethan Harvey, Mikhail Petrov, and Michael~C. Hughes.
\newblock {Transfer Learning with Informative Priors: Simple Baselines Better than Previously Reported}.
\newblock \emph{Transactions on Machine Learning Research (TMLR)}, 2024.
\newblock ISSN 2835-8856.
\newblock URL \url{https://openreview.net/forum?id=BbvSU02jLg}.
\newblock Reproducibility Certification.

\bibitem[Hastie et~al.(2009)Hastie, Tibshirani, and Friedman]{hastie2009elements}
Trevor Hastie, Robert Tibshirani, and Jerome Friedman.
\newblock \emph{{The Elements of Statistical Learning}}, chapter Sec. 3.4 Shrinkage Methods.
\newblock Springer Series in Statistics. Springer, second edition, 2009.
\newblock URL \url{https://web.stanford.edu/~hastie/ElemStatLearn/}.

\bibitem[He et~al.(2016)He, Zhang, Ren, and Sun]{he2016deep}
Kaiming He, Xiangyu Zhang, Shaoqing Ren, and Jian Sun.
\newblock {Deep Residual Learning for Image Recognition}.
\newblock In \emph{Proceedings of the IEEE/CVF Conference on Computer Vision and Pattern Recognition (CVPR)}, 2016.

\bibitem[Higgins et~al.(2017)Higgins, Matthey, Pal, Burgess, Glorot, Botvinick, Mohamed, and Lerchner]{higginsBetaVAE2017}
Irina Higgins, Loic Matthey, Arka Pal, Christopher Burgess, Xavier Glorot, Matthew Botvinick, Shakir Mohamed, and Alexander Lerchner.
\newblock {$\beta$-VAE: Learning Basic Visual Concepts with a Constrained Variational Framework}.
\newblock In \emph{International Conference on Learning Representations (ICLR)}, 2017.

\bibitem[Hinton and Van~Camp(1993)]{hinton1993keeping}
Geoffrey~E. Hinton and Drew Van~Camp.
\newblock {Keeping Neural Networks Simple by Minimizing the Description Length of the Weights}.
\newblock In \emph{Proceedings of the sixth annual conference on Computational Learning Theory}, 1993.

\bibitem[Jeffreys(1939)]{jeffreys1939theory}
Harold Jeffreys.
\newblock \emph{The Theory of Probability}.
\newblock The Clarendon Press, Oxford, 1939.

\bibitem[Jordan et~al.(1999)Jordan, Ghahramani, Jaakkola, and Saul]{jordan1999introduction}
Michael~I Jordan, Zoubin Ghahramani, Tommi~S Jaakkola, and Lawrence~K Saul.
\newblock An introduction to variational methods for graphical models.
\newblock \emph{Machine learning}, 37:\penalty0 183--233, 1999.

\bibitem[Kapoor et~al.(2022)Kapoor, Maddox, Izmailov, and Wilson]{kapoor2022uncertainty}
Sanyam Kapoor, Wesley~J. Maddox, Pavel Izmailov, and Andrew~G. Wilson.
\newblock {On Uncertainty, Tempering, and Data Augmentation in Bayesian Classification}.
\newblock In \emph{Advances in Neural Information Processing Systems (NeurIPS)}, 2022.

\bibitem[Kobak et~al.(2020)Kobak, Lomond, and Sanchez]{kobak2020optimal}
Dmitry Kobak, Jonathan Lomond, and Benoit Sanchez.
\newblock {The Optimal Ridge Penalty for Real-world High-dimensional Data Can Be Zero or Negative due to the Implicit Ridge Regularization}.
\newblock \emph{Journal of Machine Learning Research (JMLR)}, 21\penalty0 (169):\penalty0 1--16, 2020.

\bibitem[Krishnan et~al.(2019)Krishnan, Subedar, and Tickoo]{krishnan2019efficient}
Ranganath Krishnan, Mahesh Subedar, and Omesh Tickoo.
\newblock {Efficient Priors for Scalable Variational Inference in Bayesian Deep Neural Networks}.
\newblock In \emph{Proceedings of the IEEE/CVF International Conference on Computer Vision Workshops (ICCVW)}, 2019.

\bibitem[Krizhevsky et~al.(2010)Krizhevsky, Nair, and Hinton]{krizhevsky2010cifar}
Alex Krizhevsky, Vinod Nair, and Geoffrey Hinton.
\newblock {CIFAR-10 (Canadian Institute for Advanced Research)}.
\newblock 2010.
\newblock URL \url{http://www.cs.toronto.edu/~kriz/cifar.html}.

\bibitem[Krogh and Hertz(1991)]{krogh1991simple}
Anders Krogh and John~A. Hertz.
\newblock {A Simple Weight Decay Can Improve Generalization}.
\newblock In \emph{Advances in Neural Information Processing Systems (NeurIPS)}, 1991.

\bibitem[Loshchilov and Hutter(2016)]{loshchilov2016sgdr}
Ilya Loshchilov and Frank Hutter.
\newblock {SGDR: Stochastic Gradient Descent with Warm Restarts}.
\newblock In \emph{Proceedings of the International Conference on Learning Representations (ICLR)}, 2016.

\bibitem[Lotfi et~al.(2022)Lotfi, Izmailov, Benton, Goldblum, and Wilson]{lotfi2022bayesian}
Sanae Lotfi, Pavel Izmailov, Gregory Benton, Micah Goldblum, and Andrew~Gordon Wilson.
\newblock {Bayesian Model Selection, the Marginal Likelihood, and Generalization}.
\newblock In \emph{International Conference on Machine Learning (ICML)}, 2022.

\bibitem[MacKay(1991)]{mackay1991bayesian}
David MacKay.
\newblock {Bayesian Model Comparison and Backprop Nets}.
\newblock In \emph{Advances in Neural Information Processing Systems (NeurIPS)}, 1991.

\bibitem[Maddox et~al.(2019)Maddox, Izmailov, Garipov, Vetrov, and Wilson]{maddox2019simple}
Wesley~J Maddox, Pavel Izmailov, Timur Garipov, Dmitry~P Vetrov, and Andrew~Gordon Wilson.
\newblock {A Simple Baseline for Bayesian Uncertainty in Deep Learning}.
\newblock In \emph{Advances in Neural Information Processing Systems (NeurIPS)}, 2019.

\bibitem[Miller and Dunson(2019)]{miller2019robust}
Jeffrey~W. Miller and David~B. Dunson.
\newblock {Robust Bayesian Inference via Coarsening}.
\newblock \emph{Journal of the American Statistical Association}, 2019.
\newblock URL \url{http://arxiv.org/abs/1506.06101}.

\bibitem[Mohamed et~al.(2020)Mohamed, Rosca, Figurnov, and Mnih]{mohamed2020monte}
Shakir Mohamed, Mihaela Rosca, Michael Figurnov, and Andriy Mnih.
\newblock {Monte Carlo Gradient Estimation in Machine Learning}.
\newblock \emph{Journal of Machine Learning Research (JMLR)}, 21\penalty0 (132), 2020.
\newblock URL \url{http://jmlr.org/papers/v21/19-346.html}.

\bibitem[Murphy(2022{\natexlab{a}})]{murphy2022Example}
Kevin~S. Murphy.
\newblock \emph{Probabilistic Machine Learning: An Introduction}, chapter 6.2.3 Example: KL divergence between two Gaussians.
\newblock MIT Press, 2022{\natexlab{a}}.

\bibitem[Murphy(2022{\natexlab{b}})]{murphy2022Picking}
Kevin~S. Murphy.
\newblock \emph{Probabilistic Machine Learning: An Introduction}, chapter 4.5.4 Picking the regularizer using a validation set.
\newblock MIT Press, 2022{\natexlab{b}}.

\bibitem[Murphy(2022{\natexlab{c}})]{murphy2022Regularization}
Kevin~S. Murphy.
\newblock \emph{Probabilistic Machine Learning: An Introduction}, chapter 4.5 Regularization.
\newblock MIT Press, 2022{\natexlab{c}}.

\bibitem[Nilsback and Zisserman(2008)]{nilsback2008automated}
Maria-Elena Nilsback and Andrew Zisserman.
\newblock {Automated Flower Classification over a Large Number of Classes}.
\newblock In \emph{2008 Sixth Indian Conference on Computer Vision, Graphics \& Image Processing (ICVGIP)}, 2008.

\bibitem[Parkhi et~al.(2012)Parkhi, Vedaldi, Zisserman, and Jawahar]{parkhi2012cats}
Omkar~M Parkhi, Andrea Vedaldi, Andrew Zisserman, and CV~Jawahar.
\newblock {Cats And Dogs}.
\newblock In \emph{Proceedings of the IEEE/CVF Conference on Computer Vision and Pattern Recognition (CVPR)}, 2012.

\bibitem[Raschka(2018)]{raschka2018model}
Sebastian Raschka.
\newblock {Model Evaluation, Model Selection, and Algorithm Selection in Machine Learning}.
\newblock \emph{arXiv preprint arXiv:1811.12808}, 2018.
\newblock URL \url{http://arxiv.org/abs/1811.12808}.

\bibitem[Rudner et~al.(2024)Rudner, Pan, Li, Shwartz-Ziv, and Wilson]{rudner2024finetuning}
Tim G.~J. Rudner, Xiang Pan, Yucen~Lily Li, Ravid Shwartz-Ziv, and Andrew~Gordon Wilson.
\newblock {Fine-Tuning with Uncertainty-Aware Priors Makes Vision and Language Foundation Models More Reliable}.
\newblock In \emph{ICML Workshop on Structured Probabilistic Inference {\&} Generative Modeling (SPIGM@ICML)}, 2024.
\newblock URL \url{https://openreview.net/forum?id=37fM2QEBSE}.

\bibitem[Sharif~Razavian et~al.(2014)Sharif~Razavian, Azizpour, Sullivan, and Carlsson]{sharif2014cnn}
Ali Sharif~Razavian, Hossein Azizpour, Josephine Sullivan, and Stefan Carlsson.
\newblock {CNN Features off-the-shelf: an Astounding Baseline for Recognition}.
\newblock In \emph{Proceedings of the IEEE Conference on Computer Vision and Pattern Recognition Workshops (CVPRW)}, 2014.

\bibitem[Shwartz-Ziv et~al.(2022)Shwartz-Ziv, Goldblum, Souri, Kapoor, Zhu, LeCun, and Wilson]{shwartz2022pre}
Ravid Shwartz-Ziv, Micah Goldblum, Hossein Souri, Sanyam Kapoor, Chen Zhu, Yann LeCun, and Andrew~G. Wilson.
\newblock {Pre-Train Your Loss: Easy Bayesian Transfer Learning with Informative Priors}.
\newblock In \emph{Advances in Neural Information Processing Systems (NeurIPS)}, 2022.
\newblock URL \url{https://proceedings.neurips.cc/paper_files/paper/2022/file/b1e7f61f40d68b2177857bfcb195a507-Paper-Conference.pdf}.

\bibitem[{\v{S}}pendl and Pirc(2023)]{spendl2023easy}
Martin {\v{S}}pendl and Klementina Pirc.
\newblock {[Re] Easy Bayesian Transfer Learning with Informative Priors}.
\newblock In \emph{ML Reproducibility Challenge 2022}, 2023.
\newblock URL \url{https://openreview.net/forum?id=JpaQ8GFOVu}.

\bibitem[Wang et~al.(2017)Wang, Kucukelbir, and Blei]{wang2017robust}
Yixin Wang, Alp Kucukelbir, and David~M Blei.
\newblock {Robust Probabilistic Modeling with Bayesian Data Reweighting}.
\newblock In \emph{International Conference on Machine Learning (ICML)}, 2017.
\newblock URL \url{https://proceedings.mlr.press/v70/wang17g/wang17g.pdf}.

\bibitem[Wenzel et~al.(2020)Wenzel, Roth, Veeling, {\'S}wi{\k{a}}tkowski, Tran, Mandt, Snoek, Salimans, Jenatton, and Nowozin]{wenzel2020good}
Florian Wenzel, Kevin Roth, Bastiaan~S. Veeling, Jakub {\'S}wi{\k{a}}tkowski, Linh Tran, Stephan Mandt, Jasper Snoek, Tim Salimans, Rodolphe Jenatton, and Sebastian Nowozin.
\newblock {How Good is the Bayes Posterior in Deep Neural Networks Really?}
\newblock In \emph{International Conference on Machine Learning (ICML)}, 2020.

\bibitem[Woodbury(1950)]{woodbury1950inverting}
Max~A. Woodbury.
\newblock \emph{Inverting Modified Matrices}.
\newblock Department of Statistics, Princeton University, 1950.

\bibitem[Xu et~al.(2019)Xu, Quiroz, Kohn, and Sisson]{xu2019variance}
Ming Xu, Matias Quiroz, Robert Kohn, and Scott~A. Sisson.
\newblock {Variance Reduction Properties of the Reparameterization Trick}.
\newblock In \emph{International Conference on Artificial Intelligence and Statistics (AISTATS)}, 2019.

\bibitem[Xuhong et~al.(2018)Xuhong, Grandvalet, and Davoine]{xuhong2018explicit}
Li~Xuhong, Yves Grandvalet, and Franck Davoine.
\newblock {Explicit Inductive Bias for Transfer Learning with Convolutional Networks}.
\newblock In \emph{International Conference on Machine Learning (ICML)}, 2018.

\bibitem[Zhang and Kjellstr{\"o}m(2014)]{zhang2015supervise}
Cheng Zhang and Hedvig Kjellstr{\"o}m.
\newblock How to supervise topic models.
\newblock In \emph{ECCV Workshop on Graphical Models in Computer Vision}, 2014.

\end{thebibliography}


\appendix

\section{Classification}
\label{sec:classification}

\subsection{Dataset details}
\label{sec:dataset_details}
We inlcude experiments on CIFAR-10 \citep{krizhevsky2010cifar} and Oxford-IIIT Pet \citep{parkhi2012cats}.
For both datasets, we enforce classes are uniformly distributed in the training data.

We use the same preprocessing steps for both datasets.
For each distinct training set size $N$, we compute the mean and standard deviation of each channel to normalize images.
During fine-tuning we resize the images to $256 \times 256$ pixels, perform random cropping to $224 \times 224$, and perform horizontal flips.
At test time, we resize the images to $256 \times 256$ pixels and center crop to $224 \times 224$.

\subsection{Classifier details}
\label{sec:classifier_details}

We use SGD with a Nesterov momentum parameter of 0.9 and batch size of 128 for optimization.
We train for 6,000 steps using a cosine annealing learning rate \citep{loshchilov2016sgdr}.

For \baseline, we select the initial learning rate from $\{0.1, 0.01, 0.001, 0.0001\}$. 
For L2-zero and L2-SP we select $\frac{\alpha}{N}, \frac{\beta}{N}$ from $\{0.01, 0.001, 0.0001, 1\text{e-}5, 1\text{e-}6, 0.0\}$.
For PTYL, we select $\lambda$ from 10 logarithmically spaced values between 1e0 to 1e9 and $\frac{1}{\tau N}$ from $\{0.01, 0.001, 0.0001, 1\text{e-}5, 1\text{e-}6, 0.0\}$.

While tuning hyperparameters, we hold out 1/5 of the training set for validation, ensuring balanced class frequencies between sets.

After selecting the optimal hyperparameters from the validation set NLL, we retrain the model using the selected hyperparameters on the combined set of all $N$ images (merging training and validation). 
All results report performance on the task in question's predefined test set.

For DE ELBo, we select the initial learning rate from $\{0.1, 0.01, 0.001, 0.0001\}$ with the highest training set $\JELBO$.
For hyperparameter/model selection (the last epoch), we average over 10 samples to get a Monte Carlo approximation of the expectation.

\section{Learning Key Hyperparameters $\lambda, \tau$ with the ELBo}
\label{sec:learning_key_hyperparameters}
In our particular model in Eq.~(\ref{eq:joint_pdf_model}), the KL divergence between two Gaussians \citep{murphy2022Example} simplifies for the classifier head KL term as:
\begin{align}
    -\mathbb{KL}(q(V) \| p(V)) = -\frac{1}{2} \left[ \frac{\bar{\sigma}^2}{\tau} D + \frac{1}{\tau} || \text{vec}(\bar{V}) ||_2^2 - D + \log \left( \frac{\tau^{D}}{\bar{\sigma}^{2D}} \right)\right].
\end{align}
\paragraph{Closed-form updates}
To find an optimal $\tau$ value with respect to the $\JELBO$, notice that of the 3 additive terms in Eq.~\eqref{eq:elbo_objective}, only the KL term between $q(V)$ and $p(V)$ involves $\tau$. We solve for $\tau$ by taking the gradient of the KL term with respect to $\tau$, setting to zero, and solving, with assurances of a local maximum of $\JELBO$ via a second derivative test (see App.~\ref{sec:second_derivative_test}). The gradient is
\begin{align}
    \nabla_\tau -\mathbb{KL}(q(V) \| p(V)) = -\frac{1}{2} \left[ - \frac{\bar{\sigma}^2}{\tau^2} D - \frac{1}{\tau^2} || \text{vec}(\bar{V}) ||_2^2 + \frac{D}{\tau} \right].
\end{align}
Setting $\nabla_\tau -\mathbb{KL}(q(V) \| p(V)) = 0$ and solving for $\tau$, we get 
\begin{align}
    \tau^* = \bar{\sigma}^2 + \frac{1}{D} || \text{vec}(\bar{V}) ||_2^2.
\end{align}

\section{Second Derivative Test}
\label{sec:second_derivative_test}

The second derivative is
\begin{align}
    \nabla^2_\lambda -\mathbb{KL}(q(w) \| p(w)) &= -\frac{1}{2} \left[ \frac{2\bar{\sigma}^2}{\lambda^3} \Tr(\Sigma_p^{-1}) + \frac{2}{\lambda^3} (\mu_p - \bar{w})^T \Sigma_p^{-1} (\mu_p - \bar{w}) - \frac{D}{\lambda^2} \right] \\
    &= -\frac{1}{2} \left[ \frac{2D}{\lambda^3} \frac{1}{D} \left( \bar{\sigma}^2 \Tr(\Sigma_p^{-1}) + (\mu_p - \bar{w})^T \Sigma_p^{-1} (\mu_p - \bar{w}) \right) - \frac{D}{\lambda^2} \right] \\
    &= -\frac{1}{2} \left[ \frac{2D}{\lambda^3} \lambda^* - \frac{D}{\lambda^2} \right].
\end{align}
Plugging in $\lambda^*$ and simplifying, we get
\begin{align}
    \nabla^2_\lambda -\mathbb{KL}(q(w) \| p(w | \lambda^*)) &= -\frac{D}{2}  \frac{1}{\lambda^{*2}}
\end{align}
This expression is always negative, indicating that $\lambda^*$ is a local maximum of $\JELBO$.

The second derivative is
\begin{align}
    \nabla_\tau^2 -\mathbb{KL}(q(V) \| p(V)) &= -\frac{1}{2} \left[ \frac{2\bar{\sigma}^2}{\tau^3} D + \frac{2}{\tau^3} || \text{vec}(\bar{V}) ||_2^2 - \frac{D}{\tau^2} \right] \\
    &= -\frac{1}{2} \left[  \frac{2D}{\tau^3}  \left( \bar{\sigma}^2 + \frac{1}{D} || \text{vec}(\bar{V}) ||_2^2 \right) - \frac{D}{\tau^2} \right] \\ 
    &= -\frac{1}{2} \left[  \frac{2D}{\tau^3} \tau^*  - \frac{D}{\tau^2} \right].
\end{align}
Plugging in $\tau^*$ and simplifying, we get
\begin{align}
    \nabla^2_\tau -\mathbb{KL}(q(V) \| p(V | \tau^*)) &= -\frac{D}{2}  \frac{1}{\tau^{*2}}.
\end{align}
This expression is always negative, indicating that $\tau^*$ is a local maximum of $\JELBO$.

\section{Low-Rank $\Sigma_p$}
\label{sec:low-rank}
The PTYL method \citep{shwartz2022pre} uses Stochastic Weight Averaging-Gaussian (SWAG) \citep{maddox2019simple} to approximate the posterior distribution $p(w|\mathcal{D}_S)$ of the source data $\mathcal{D}_S$ with a Gaussian distribution $\mathcal{N}(\mu, \Sigma)$ where $\mu$ is the learned mean and $\Sigma = \frac{1}{2}(\Sigma_{\text{diag}} + \Sigma_{\text{LR}})$ is a representation of a covariance matrix with both diagonal and \emph{low-rank} components.
The LR covariance has the form $\Sigma_{\textrm{LR}} = \frac{1}{K-1} Q Q^T$, where $Q \in \mathbb{R}^{D \times K}$.

We use the Woodbury matrix identity \cite{woodbury1950inverting}, trace properties, and the matrix determinant lemma to compute the trace of the inverse, squared Mahalanobis distance, and log determinant of the low-rank covariance matrix for the KL term.

The trace and log determinant of the low-rank covariance matrix can be calculated once and used during training.
Just like in PTYL method, the squared Mahalanobis distance needs to be re-evaluated every iteration of gradient descent.

\subsection{Trace of the inverse}
We compute the trace of the inverse of the low-rank covariance matrix using the Woodbury matrix identity and trace properties.
\begin{align*}
\Tr (\Sigma_p^{-1}) &= \Tr( (A + UCV )^{-1} ) \\
&= \Tr (A^{-1} - A^{-1}U(C^{-1} + VA^{-1}U)^{-1}VA^{-1}) && \text{Woodbury matrix identity} \\
&= \Tr (A^{-1}) - \Tr(A^{-1}U(C^{-1} + VA^{-1}U)^{-1}VA^{-1}) && \Tr(A+B) = \Tr(A) + \Tr(B) \\
&= \Tr (A^{-1}) - \Tr((C^{-1} + VA^{-1}U)^{-1}VA^{-1}A^{-1}U) && \Tr(AB) = \Tr(BA)
\end{align*}
where $A=\frac{1}{2}\Sigma_{\text{diag}}$, $C = I_K$, $U = \frac{1}{\sqrt{2K-2}}Q$, and $V=\frac{1}{\sqrt{2K-2}}Q^T$.
The last trace property, lets us compute the trace of the inverse of the low-rank covariance matrix without having to store a $D \times D$ covariance matrix.

\subsection{Squared Mahalanobis distance}
We compute the squared Mahalanobis distance $(\mu_p-\bar{w})^T\Sigma_p^{-1}(\mu_p-\bar{w})$ by distributing the mean difference vector into the Woodbury matrix identity.
\begin{align*}
\Sigma_p^{-1} &= (A + UCV )^{-1} \\
&= (A^{-1} - A^{-1}U(C^{-1} + VA^{-1}U)^{-1}VA^{-1}) && \text{Woodbury matrix identity}
\end{align*}

\subsection{Log determinant}
We compute the log determinant of the low-rank covariance matrix using the matrix determinant lemma.
\begin{align*}
\log \det(\Sigma_p) &= \log \det(A + UV) \\
&= \log (\det(I_K + VA^{-1}U) \det(A))  && \text{Matrix determinant lemma}
\end{align*}

\section{NLL Results}
\label{sec:nll_results}
\setlength{\tabcolsep}{4pt}
\begin{table}[htbp!]
  \caption{NLL on CIFAR-10 test set for different probabilistic models and methods. We report mean (min-max) over 3 separately-sampled training sets. For each separately-sampled training set and training set size, the \emph{\baselineLong} baseline requires 24 different SGD runs for L2-zero and L2-SP, and 240 for PTYL. Our \emph{data-emphasized ELBo} (DE ELBo) requires 4 different SGD runs (one for each initial learning rate) and learns optimal $\lambda,\tau$ values. See App.~\ref{sec:classifier_details} for hyperparameter search space details.}
  \label{tab:CIFAR-10_nll}
  \centering
  \footnotesize
  \begin{tabular}{llcccc}
    \toprule
    \bfseries Model & \bfseries Method & $N =$ {\bfseries 100 (10/cl.)} & \bfseries 1000 (100/cl.) & \bfseries 10000 (1k/cl.) & \bfseries 50000 (5k/cl.) \\
    \midrule
    \rowcolor{bright-gray} L2-zero & \baseline & 1.03 (0.96-1.10) & 0.42 (0.37-0.51) & 0.18 (0.16-0.19) & 0.10 (0.10-0.10) \\
    & DE ELBo & 2.99 (2.67-3.17) & 0.54 (0.53-0.54) & 0.27 (0.25-0.29) & 0.27 (0.26-0.29) \\
    \rowcolor{bright-gray} L2-SP & \baseline & 0.97 (0.94-1.01) & 0.40 (0.38-0.44) & 0.15 (0.14-0.16) & 0.09 (0.09-0.09) \\
    & DE ELBo & 1.07 (0.97-1.17) & 0.40 (0.39-0.42) & 0.26 (0.25-0.27) & 0.12 (0.11-0.12) \\   
    \rowcolor{bright-gray} PTYL & \baseline & 0.98 (0.95-1.03) & 0.41 (0.37-0.46) & 0.16 (0.16-0.17) & 0.09 (0.09-0.09) \\ 
    & DE ELBo & 1.06 (0.98-1.17) & 0.40 (0.39-0.42) & 0.26 (0.25-0.28) & 0.12 (0.11-0.12) \\
    \bottomrule
  \end{tabular}
\end{table}
\setlength{\tabcolsep}{6pt}

\setlength{\tabcolsep}{4pt}
\begin{table}[htbp!]
  \caption{NLL on Oxford-IIIT Pet test set for different probabilistic models and methods. We report mean (min-max) over 3 separately-sampled training sets. For each separately-sampled training set and training set size, the \emph{\baselineLong} baseline requires 24 different SGD runs for L2-zero and L2-SP, and 240 for PTYL. Our \emph{data-emphasized ELBo} (DE ELBo) requires 4 different SGD runs (one for each initial learning rate) and learns optimal $\lambda,\tau$ values. See App.~\ref{sec:classifier_details} for hyperparameter search space details.}
  \label{tab:Oxford-IIIT_Pet_nll}
  \centering
  \footnotesize
  \begin{tabular}{llcccc}
    \toprule
    \bfseries Model & \bfseries Method & $N~{=}$ {\bfseries 370 (10/cl.)} & \bfseries 3441(93/cl.) \\
    \midrule
    \rowcolor{bright-gray} L2-zero & \baseline & 0.44 (0.42-0.48) & 0.25 (0.24-0.26) \\
    & DE ELBo & 0.91 (0.84-1.01) & 0.27 (0.26-0.28) \\
    \rowcolor{bright-gray} L2-SP & \baseline & 0.44 (0.42-0.48) & 0.24 (0.24-0.25)\\
    & DE ELBo & 0.57 (0.55-0.61) & 0.24 (0.24-0.24) \\
    \rowcolor{bright-gray} PTYL & \baseline & 0.45 (0.42-0.48) & 0.29 (0.26-0.34) \\
    & DE ELBo & 0.57 (0.55-0.61) & 0.24 (0.24-0.24) \\
    \bottomrule
  \end{tabular}
\end{table}
\setlength{\tabcolsep}{6pt}

\newpage
\section{Supervised or Self-Supervised Priors}
\label{sec:supervised_or_self-supervised_priors}
\setlength{\tabcolsep}{4pt}
\begin{table}[htbp!]
  \caption{Accuracy on CIFAR-10 test set for different probabilistic models and methods. We report mean (min-max) over 3 separately-sampled training sets.}
  \label{tab:PTYL_CIFAR-10_acc}
  \centering
  \footnotesize
  \begin{tabular}{llcccc}
    \toprule
    \bfseries Model & \bfseries Method & $N =$ {\bfseries 100 (10/cl.)} & \bfseries 1000 (100/cl.) & \bfseries 10000 (1k/cl.) & \bfseries 50000 (5k/cl.) \\
    \midrule
    \rowcolor{bright-gray} PTYL & \baseline & 67.5 (65.7-68.4) & 87.9 (86.9-89.2) & 95.2 (95.0-95.4) & 97.3 (97.3-97.3) \\
    \rowcolor{bright-gray} PTYL (SSL) & \baseline & 58.7 (56.4-60.6) & 83.5 (83.3-83.9) & 93.8 (93.4-94.1) & 96.9 (96.7-97.0) \\
    \bottomrule
  \end{tabular}
\end{table}
\setlength{\tabcolsep}{6pt}

\setlength{\tabcolsep}{4pt}
\begin{table}[htbp!]
  \caption{NLL on CIFAR-10 test set for different probabilistic models and methods. We report mean (min-max) over 3 separately-sampled training sets.}
  \label{tab:PTYL_CIFAR-10_nll}
  \centering
  \footnotesize
  \begin{tabular}{llcccc}
    \toprule
    \bfseries Model & \bfseries Method & $N =$ {\bfseries 100 (10/cl.)} & \bfseries 1000 (100/cl.) & \bfseries 10000 (1k/cl.) & \bfseries 50000 (5k/cl.) \\
    \midrule
    \rowcolor{bright-gray} PTYL & \baseline & 0.98 (0.95-1.03) & 0.41 (0.37-0.46) & 0.16 (0.16-0.17) & 0.09 (0.09-0.09) \\ 
    \rowcolor{bright-gray} PTYL (SSL) & \baseline & 1.31 (1.20-1.44) & 0.58 (0.56-0.58) & 0.22 (0.21-0.23) & 0.10 (0.10-0.11) \\
    \bottomrule
  \end{tabular}
\end{table}
\setlength{\tabcolsep}{6pt}

\setlength{\tabcolsep}{4pt}
\begin{table}[htbp!]
  \caption{Accuracy on Oxford-IIIT Pet test set for different probabilistic models and methods. We report mean (min-max) over 3 separately-sampled training sets.}
  \label{tab:PTYL_Oxford-IIIT_Pet_acc}
  \centering
  \footnotesize
  \begin{tabular}{llcccc}
    \toprule
    \bfseries Model & \bfseries Method & $N~{=}$ {\bfseries 370 (10/cl.)} & \bfseries 3441(93/cl.) \\
    \midrule
    \rowcolor{bright-gray} PTYL & \baseline & 87.1 (86.6-87.5) & 92.2 (90.7-93.2) \\ 
    \rowcolor{bright-gray} PTYL (SSL) & \baseline & 57.4 (56.2-58.2) & 86.7 (85.0-87.8) \\
    \bottomrule
  \end{tabular}
\end{table}
\setlength{\tabcolsep}{6pt}

\setlength{\tabcolsep}{4pt}
\begin{table}[htbp!]
  \caption{NLL on Oxford-IIIT Pet test set for different probabilistic models and methods. We report mean (min-max) over 3 separately-sampled training sets.}
  \label{tab:PTYL_Oxford-IIIT_Pet_nll}
  \centering
  \footnotesize
  \begin{tabular}{llcccc}
    \toprule
    \bfseries Model & \bfseries Method & $N~{=}$ {\bfseries 370 (10/cl.)} & \bfseries 3441(93/cl.) \\
    \midrule
    \rowcolor{bright-gray} PTYL & \baseline & 0.45 (0.42-0.48) & 0.29 (0.26-0.34) \\
    \rowcolor{bright-gray} PTYL (SSL) & \baseline & 1.69 (1.64-1.72) & 0.52 (0.52-0.53) \\    
    \bottomrule
  \end{tabular}
\end{table}
\setlength{\tabcolsep}{6pt}


\end{document}